\documentclass[letterpaper, 10 pt, conference]{ieeeconf}  

\IEEEoverridecommandlockouts                              

\usepackage{graphics} 
\usepackage{epsfig} 
\usepackage{mathptmx} 
\usepackage{times} 
\usepackage{amsmath} 
\usepackage{amssymb}  
\usepackage{multirow}
\usepackage{color,soul}
\usepackage{algorithm}
\usepackage{siunitx}
\usepackage{algpseudocode}

\usepackage{hyperref}
\usepackage[comma,numbers]{natbib}

\title{\LARGE \bf
Footstep Planning for Autonomous Walking Over Rough Terrain
}

\author{Robert J. Griffin, Georg Wiedebach, Stephen McCrory, Sylvain Bertrand, Inho Lee, Jerry Pratt
\thanks{This work was funded through the ONR Grant N00014-19-1-2023 and by NASA Grant 80NSSC18M0071.}
\thanks{The authors are with the Florida Institute for Human and Machine Cognition, 40 S Alcaniz St, Pensacola, FL 32502, United States}
\thanks{Email : \url{ {rgriffin, gwiedebach, smccrory, sbertrand, ilee, jpratt}@ihmc.us}
}} 

\begin{document}

\maketitle
\thispagestyle{empty}
\pagestyle{empty}

\begin{abstract}
To increase the speed of operation and reduce operator burden, humanoid robots must be able to function autonomously, even in complex, cluttered environments.
For this to be possible, they must be able to quickly and efficiently compute desired footsteps to reach a goal.
In this work, we present a new A* footstep planner that utilizes a planar region representation of the environment enable footstep planning over rough terrain.
To increase the number of available footholds, we present an approach to allow the use of partial footholds during the planning process.
The footstep plan solutions are then post-processed to capture better solutions that lie between the lattice discretization of the footstep graph.
We then demonstrate this planner over a variety of virtual and real world environments, including some that require partial footholds and rough terrain using the Atlas and Valkyrie humanoid robots.
\end{abstract}

\section{Introduction}
\label{introduction}

One of the main developmental justifications for humanoid robots is their potential for incredible mobility.
Humans can run, swim, climb, and access arguably more environments than any other terrestrial creature of their size on the planet.
While the mobility of humanoid robots has significantly improved, allowing them to walk quickly and robustly, they are still fairly limited as to their ability to handle rough terrain.
Rough terrain for robots has a sparse, limited number of footholds, with large, discrete height changes and varied surface normals. 
This requires the robot to use accurate foot placement in order to reach the goal.
Navigating over rough terrain is a key skill for humanoid robots to function in many of their desired operating environments, and highlights their distinct mobility capabilities.
Manually placing the desired footsteps, though, is cognitively challenging and requires significant time, even for a skilled operator.
To improve the capability and usability of humanoid robots, the robot should reliably determine its own foot placement, allowing it to navigate autonomously, even in complex environments.

In this work, we focus on the development of a footstep planner with a specific focus on handling rough terrain.
The planning of feasible footholds lends itself nicely to graph-search approaches with a rich history\citep{Hornung_2012b, Hornung_2012, Chestnutt_2003, Chestnutt_2005, Chestnutt_2007, Stumpf_2014}, as considering a continuum of possible step locations is not explicitly required.
Provided a properly tuned cost-to-go heuristic, planners based on graph-search can quickly converge to a feasible solution, and lend themselves nicely to anytime approaches, providing partial solutions when there is not sufficient time to plan the entire contact sequence.
This is critical for real-world use by autonomous robots, where plans are needed on the order of fractions of seconds to prevent planning slow-downs and to adapt to dynamic environments.
Additionally, graph-search works well for nonlinear constraints and costs, where node validation is simply a binary function and cost function gradients are not used.
Lastly, graph-search is well suited to problems where the number of decision variables (in this case, footsteps) is unknown, making it very well suited to path planning over rough terrain, where the number of footsteps required can be very hard to predict.

\begin{figure}[!t]
\centering
    \includegraphics[width=0.9\columnwidth]{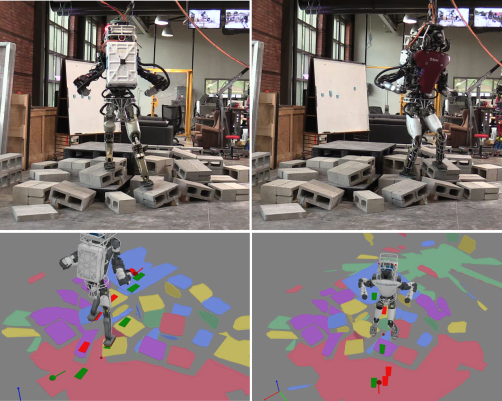}
\caption{Results of the footstep planner going up and back down a pile of cinder blocks. The pictures on bottom are screen captures of the plan from the operator interface.}
\vspace{-8mm}
\label{fig:rough_terrain_results}
\end{figure}

Here we present a novel Weighted A* footstep planner designed to handle 3D obstacles and terrain, using an efficient representation of the environment in the form of planar regions.
There have been many quite successful and thorough implementations of footstep planners in the past, including those with impressive, fielded results\citep{Deits_2014, Stumpf_2014, Karkowski_2016}.
Our focus is on the development of a planner to handle terrain with varying surface normals, large gaps, and height changes.
Uniquely, our planner allows the use of partial footholds, something we believe is required to expand the valid action set for the robot to usable size, and something that has not been demonstrated before to our knowledge.
We utilize a novel approach for pruning the possible action tree using a planar region representation of the environment.
We also present a method for post-processing the resulting plans to address cases where a more reliable foothold exists between the lattice nodes of the graph, away from edges in the world, but is typically missed due to the discretization.

To validate our planning approach, we present several experiments using the Valkyrie humanoid by NASA Johnson Space Center and the DRC Atlas by Boston Dynamics.
The experiments include walking over flat ground and avoiding dynamic obstacles, stepping stones, balance beams, and rough terrain.

\section{Related Work}

As robots have been gaining in capability, so too has the work on footstep planners been gaining interest. 
Historically, many footstep planners have used graph-search techniques.
One of the first uses of A* for footstep planning was presented in\citep{Chestnutt_2003, Chestnutt_2005}, demonstrating its viability for planning specific contact sequences for obstacle avoidance over flat ground.
The capabilities of different graph-search based planners were later compared, including an A*, ARA*, and R* planners, using several different guiding heuristics\citep{Hornung_2012}, where
it was found that R* planners are often faster, but may not produce feasible results, as expected.
More recently, a traversibility metric has been used to guide the expansion of ANA* planners to help accelerate convergence for multi-contact planning\citep{Lin_2017}.
Additionally, it has been shown that by including an estimated dynamics edge cost, the resulting plans can better adapt to the demands of different environments on the center of mass (CoM) dynamics\citep{Lin_2019}.
A planar region segmentation of the environment has also been used to extract a desired 2D body path plan for the robot, and directly used to compute the desired footholds by assuming flat ground\citep{Karkowski_2016}.
This is in contrast to the same authors' other work, which used an A* algorithm that uses an adaptive footstep expansion set to keep the search-space relatively small so that solutions are found quickly\citep{Karkowski_2016b}, using the same planar region representation as in\citep{Karkowski_2016}.

Recently searching over continuous space rather than over a discrete graph has been gaining interest. 
One of the earlier works of actually fielded continuous space algorithms used a mixed integer quadratic program (MIQP) to compute footstep plans\citep{Deits_2014}.
Like in \citep{Karkowski_2016}, this approach requires a smart segmentation of the environment into convex regions in order to be tractable.
Other planners that plan both the contact sequence and the overall motions include CHOMP \citep{Ratliff_2009}, which uses covariant gradient descent to optimize the motions and TOWR \citep{Winkler_2018}, which generates contacts and trajectories using a parameterization of end-effector and CoM motions. 
An interesting combination of the two styles is presented in\citep{Ponton_2016}, which optimizes the desired end-effector locations, including knowledge of the resulting CoM dynamics using a MIQP. 

Another common approach has been to compute footstep plans via pattern generators.
However, these applications have typically been limited to flat ground. 
One of the first examples utilized a model predictive controller to generate stable walking motions, including footsteps\citep{Herdt_2010}.
A similar approach reformulated the problem into a nonlinear optimization to include collision avoidance\citep{Naveau_2016}.
One of the most recent and promising results in using pattern generation performs multi-contact planning in two stages: first reachability planning from the body path, then planning contacts using static equilibrium, both with a bi-RRT approach, which was used to generate dynamic motions with a 3D pattern generator.
This approach was capable of generating results fast enough to plan at each robot step over short traverses\citep{Carpentier_2016}.

\section{Weighted A* Footstep Planning}
\label{framework}

The underlying algorithm comprising our footstep planner uses a Weighted A* approach\citep{Ebendt_2009}.
We break this search into the four main steps described below: node expansion, node snapping, edge checking, and edge scoring.
We also include a very brief overview of our planar region representation framework, as this is essential to the node snapping and edge checking step steps.

\subsection{Node and Graph Structure}
\label{nodeandgraphstructure}

We define each footstep node in the A* graph as a unique set of $x$, $y$, and yaw positions and robot side, making it a three dimensional structure, with edges connecting two nodes of opposite sides.
The grid is discretized into two sizes, one for $x$ and $y$, and one for yaw.
The action graph can be reduced from the $\mathbb{R}^6$ footstep pose to these three values as they fully define the pose with the foot height, pitch, and roll being constrained by the environment.
In this work, we use a $x$-$y$ grid size of \SI{5}{\cm} and a yaw size of \ang{10}.

\begin{algorithm}[b]
\caption{A* Search Algorithm}\label{alg:search_algorithm}
\begin{algorithmic}[1]
\State addStartNodeToQueue()
\While {hasNodesToCheck()}
\State nodeToExpand = getCheapestNode();
\If {hasNodeAlreadyBeenExpanded()}
\State skipToNextNodeInQueue();
\EndIf

\If {hasReachedTheGoal() or hasTimedOut()}
\State stopSearch();
\EndIf

\State childNodesToCheck = expandNode();
\For {childNode $\in$ childNodesToCheck}
\State snapChildNodeToWorld();

\State costOfEdge = getCostOfEdge();
\State addEdgeToGraph(childNode, costOfEdge);

\State costToGoal = estimateCostToGoal();
\State nodeCost = costOfPath(childNode) + costToGoal;
\State addNodeToQueue(childNode, nodeCost);
\EndFor

\EndWhile
\State footstepPlan = getBestPathToEndNode();
\end{algorithmic}
\end{algorithm}

\begin{figure}[!t]
\centering
    \includegraphics[width=0.9\columnwidth]{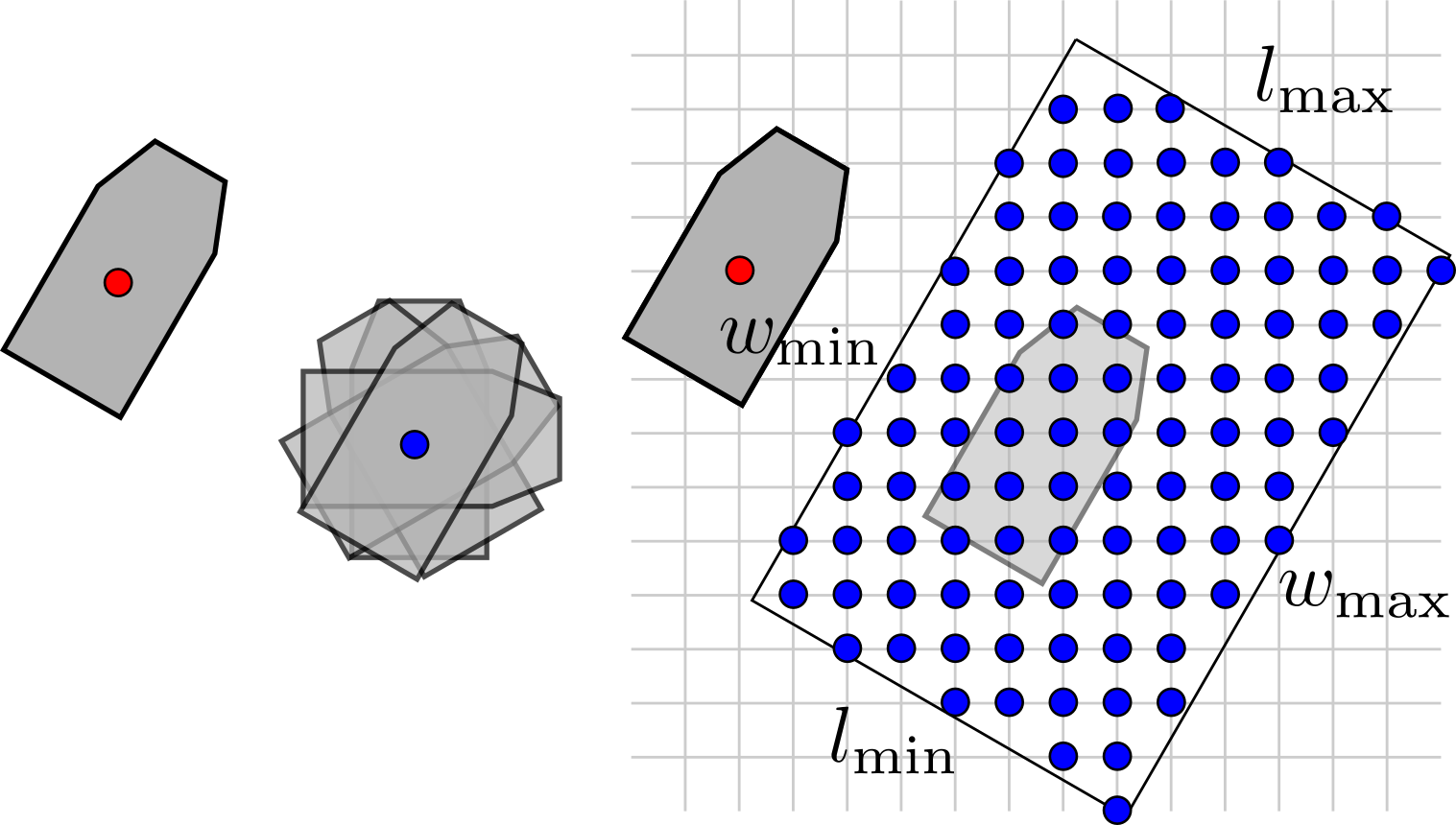}
\caption{Example of node expansion. Left: The parent node (red), with the set of all possible yaws at a single location (blue). Right: The possible expansion area is defined by maximum and minimum length and widths relative to the parent node (red). Then, a node is added at each vertex of the graph contained in this region (blue).}
\label{fig:node_expansion}
\end{figure}

\subsection{Algorithm Overview}
\label{processoverview}

All the leaves of the action tree, or ending nodes of footstep graph, are stored in a priority queue, where the node with the lowest total estimated cost is first in the queue, representing the ending nodes in the graph.
The node is then removed from the queue and expanded to determine the possible actions.
These children nodes are then checked, and if feasible, are scored and added to the graph, resolving any loop closures where necessary by picking the less expensive parent node.
If one of the children nodes equals a goal node, the algorithm is halted, as a solution has been found.
This process is described in Algorithm \autoref{alg:search_algorithm}:

\begin{figure}[!t]
\centering
    \includegraphics[width=0.9\columnwidth]{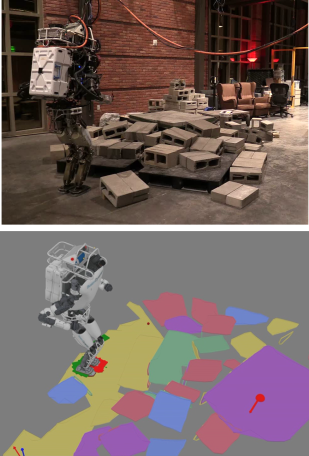}
\caption{Example of our representation of the environment as planar regions. 
Top: The robot collects a point cloud using, in this case, a spinning Hokuyo LIDAR.
Bottom: Planar surfaces are then generated from this point cloud. 
Each of these planar surfaces can be further decomposed into a set of convex regions.}
\label{fig:convex_deconstruction}
\vspace{-6mm}
\end{figure}

\subsection{Node Expansion}
\label{nodeexpansion}

The first step of every iteration is to expand the lowest cost node to determine the feasible actions, where the cost is both the cost of the path to that point, plus the estimated cost-to-go to the goal.
To determine the set of possible actions, we define a reachability box in the frame of the parent node, as shown in \autoref{fig:node_expansion}.
Then, we simply add a node to each vertex inside this reachability, shown on the right in \autoref{fig:node_expansion}, including all possible yaws at each location, as in the left of \autoref{fig:node_expansion}.
We perform a brief check on total Euclidean distance, eliminating nodes that are too far from the base node.
For the proposed experiments, this results in approximately 600 children nodes for each iteration.
For a more in-depth overview of how the node expansion step can be modified to increase the search-speed, see\citep{Karkowski_2016b}.

\subsection{Planar Region Representation}
\label{planarregionrepresentation}

One of the most important aspects for planning footsteps over complex terrain is the representation of the environment in an efficient manner for the planner.
As in\citep{Karkowski_2016, Deits_2015, Gutmann_2008}, we decompose the environment into a set of planar regions. 
Planar regions can represent a huge variety of surfaces, from estimating flat gravel as a single, large region, to a cylindrical drum as a tessellated set of regions.
Each planar surface is then deconstructed into a set of convex regions, providing a highly efficient, low order, convex representation of the world, shown in \autoref{fig:convex_deconstruction}.
We have found this to be useful and effective for footstep planning, allowing the exploitation of simple Euclidean geometry for most of the calculations.

\begin{figure}[!b]
\vspace{-4mm}
\centering
    \includegraphics[width=0.9\columnwidth]{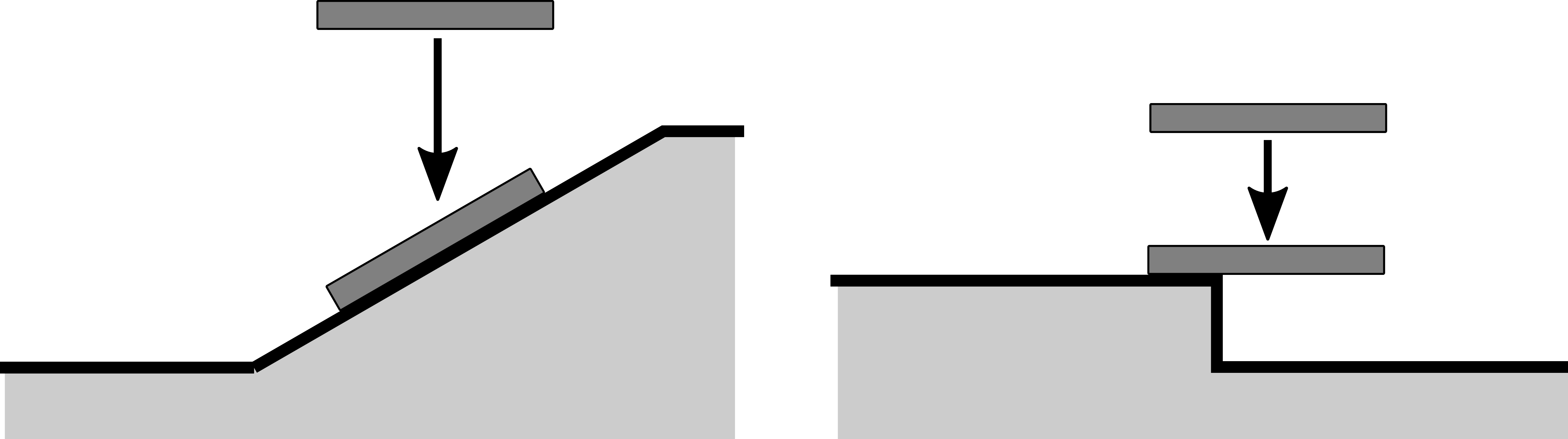}
\caption{To map the $\mathbb{R}^3$ node location to the $\mathbb{R}^6$ foot spatial location, we snap the foot polygon to the highest planar region in space.
The snap always go to the highest region.}
\label{fig:node_snapping}
\end{figure}

\begin{figure}[!t]
\centering
    \includegraphics[width=0.9\columnwidth]{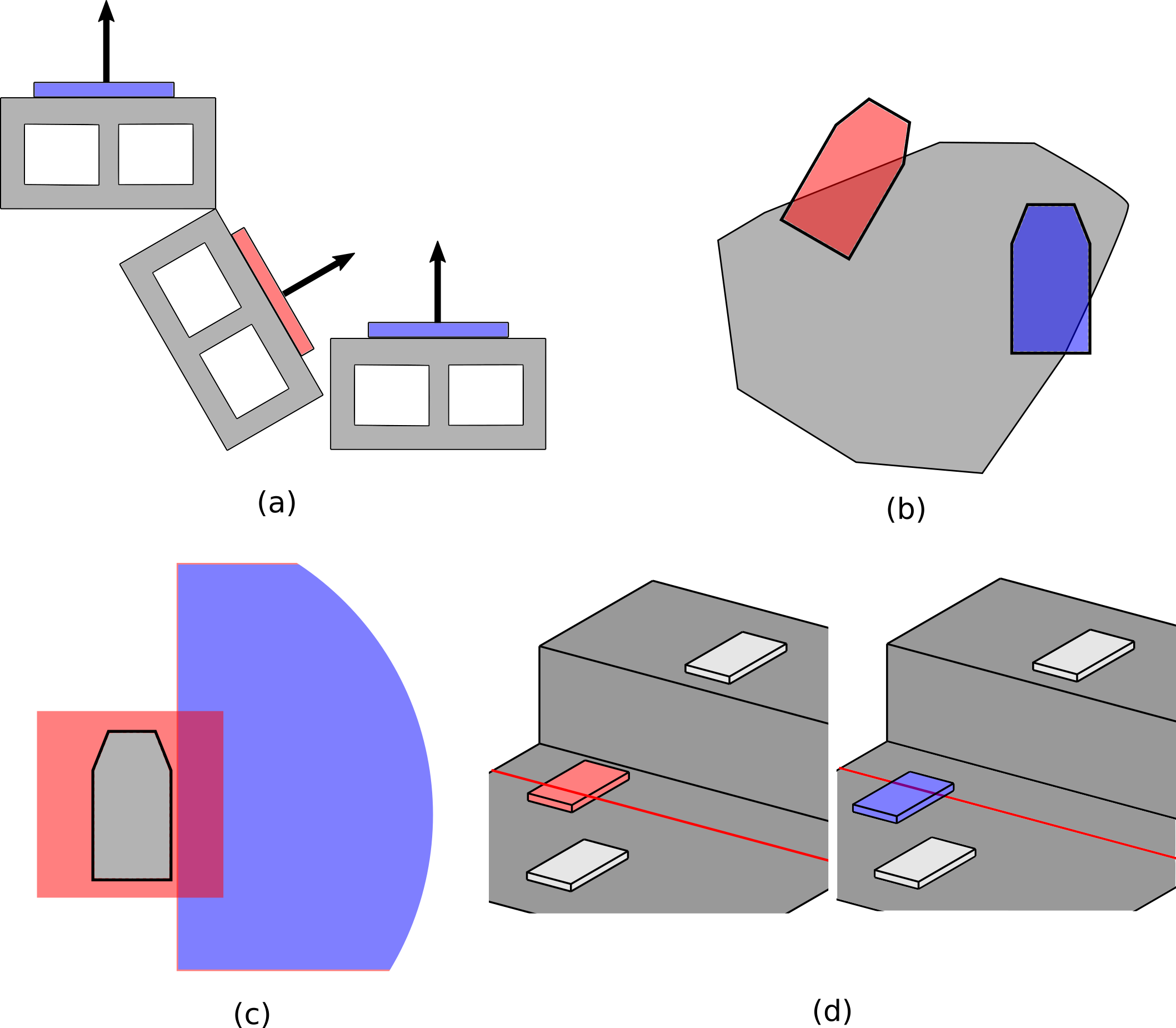}
\caption{Different edge checks, where red is rejected and blue is accepted. (a) The angle of the surface normal is evaluated to determine if the incline is too steep. (b) Nodes with too little support area are rejected, while those with enough are still accepted. 
(c) The step position is evaluated, where the red area represents a minimum clearance of the stance foot, while a max forward, backward, inside, outside and total distance is required. (d) By setting a minimum distance from the base of a cliff, shin collisions can be avoided.}
\label{fig:node_checking}
\vspace{-6mm}
\end{figure}

\subsection{Node Snapping}
\label{nodesnapping}

To evaluate the edge validity, the map of each node in the 3D footstep graph to the full 6D Euclidean footstep pose must be determined.
To do this, we simply ``snap'' the node to the planar region, as illustrated in \autoref{fig:node_snapping}.
However, as a foothold is not a simple point but is in fact a full polygon, we must snap the entire polygon to the world.
To calculate the snap, we project the foot polygon down onto each successive planar region, then keep the transform that results in the overall highest vertex in the world, as this point must be on top.

\subsection{Edge Checking}
\label{nodechecking}

To test the validity of each possible edge in the footstep graph, that is each connection between a parent and potential child node, we run through a series of checks on the child node itself and the total connection, some of which are shown in \autoref{fig:node_checking}.
First is verifying that the footstep polygon can be snapped to a planar region.
We then check that the normal of the planar region is below a certain incline, as in \autoref{fig:node_checking}(a). 
We also check that the snapped polygon contains enough intersecting area with the world, verifying that the foot has enough contact with planar region.
By allowing partial footholds we can greatly increase the number of available footholds, such as in \autoref{fig:node_checking}(b), as many nodes that are near the edge result in the foot slightly hanging off, but are still feasible.
We additionally check that the step is properly positioned with respect to the stance foot, and that the step isn't too high or too low, as shown in \autoref{fig:node_checking}(c).
For rough terrain, we also check that the foot isn't too close to the base of a cliff, or a high region in the world like a step, as shown in \autoref{fig:node_checking}(d).

\begin{figure}[!b]
\vspace{-8mm}
\centering
    \includegraphics[width=\columnwidth]{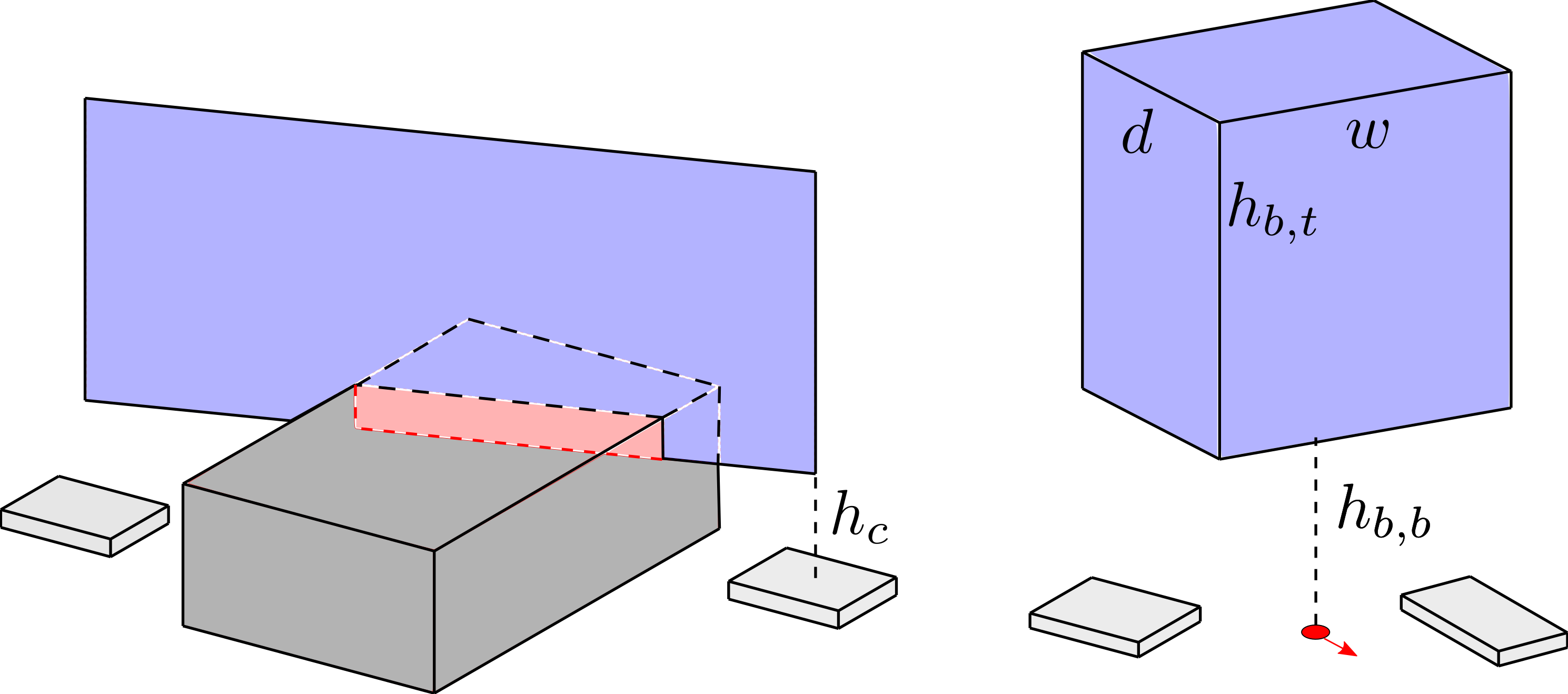}
\caption{Left: To verify that the robot isn't attempting to step over too high an obstacle, we check to make sure the plane drawn between the feet at height $h_c$ (light blue) does not collide with any obstacles in the environment, with collisions shown in red.
Right: A bounding box at the center of the midstance pose, $m$, at a bottom height $h_{b,b}$ and top height $h_{b,t}$, which is of width $w$ and depth $d$ can be checked for collisions with the environment.}
\label{fig:obstacle_collision_checker}
\end{figure}

We also verify that the robot is not attempting to step over an obstacle that is too high, as shown in \autoref{fig:obstacle_collision_checker}, left.
This is done by building a planar region that spans the two feet at a specified height.
We also build a bounding box around the midstance of the foot, similar to\citep{Deits_2014} with fewer collision boxes. 
While this one box approach is somewhat limiting, the fewer boxes greatly reduces the number of operations performed on each node in the search, which is of greater concern for graph-search type approaches than a MIQP-type approach.
This convex polytope is then checked for collisions with the world regions.

\subsection{Edge Scoring}
\label{edgescoring}

In an A* algorithm, the total cost of the path is sum of the cost of each edge along the path.
However, to help guide the expansion to increase efficiency, a cost-to-go estimate is also included, trying to estimate the total cost of the remaining distance from the current node to the goal.
The best node to expand, then, is the node with the sum of the lowest total path cost and the remaining cost estimate.
Here, we describe our path cost metrics and our cost-to-go heuristic.

\subsubsection{Step Cost}
\label{stepcost}

\begin{figure}[!t]
\centering
    \includegraphics[width=0.8\columnwidth]{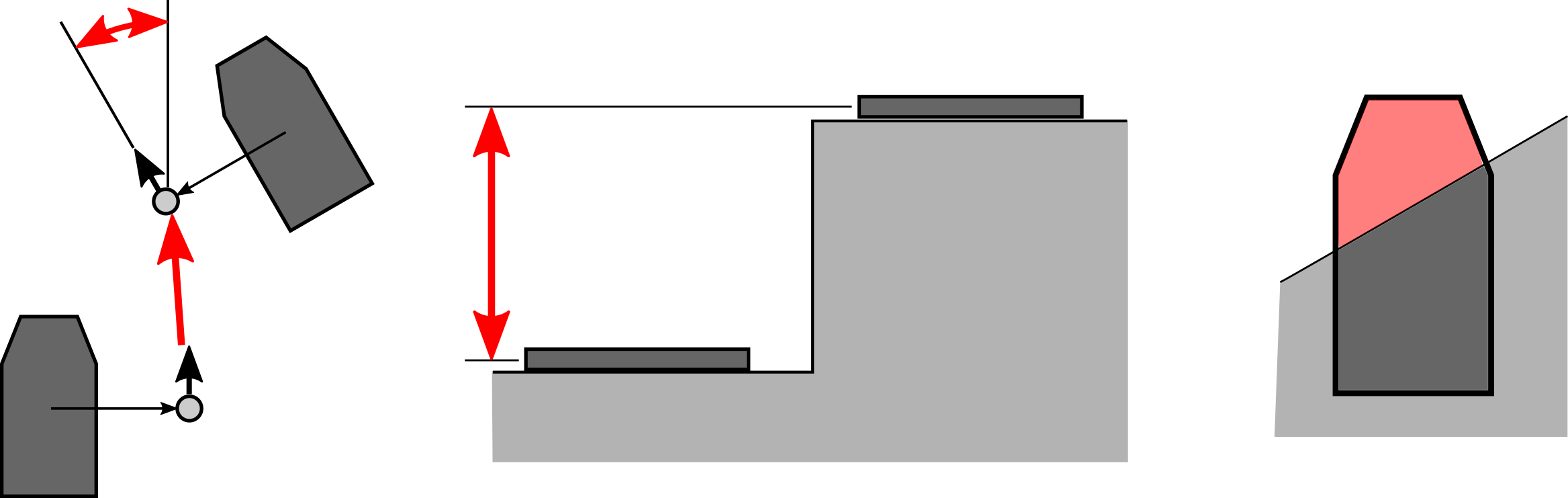}
\caption{The cost of each edge is determined by a variety of cost factors, including the translation and yaw of the midstance pose (left), the total height change of the step (middle), and the fraction of the uncovered foothold area (right).}
\vspace{-8mm}
\label{fig:edge_cost}
\end{figure}

The primary cost of the path planning problem is to minimize the total distance. 
To determine the distance cost, we use the distance between the nominal mid-stance poses, as shown in the left of \autoref{fig:edge_cost}.
We then add a cost for the height change and the yaw change of the step.
We lastly evaluate the ``quality'' of the foothold, using metrics including percent area of the foothold and the required foot roll and pitch.
To try and minimize the number of steps taken, a static cost per step is also included.

While more complex cost functions that increase the cost of yaw as the step length is increased are possible, we have not found them necessary.
The influence of each of these cost terms on the resulting path taken can be varied by altering the weight on each cost term.
Additionally, so that we can request an anytime solution, we keep track of the ongoing best path, by checking to see if the cost of each new node evaluated is less than the current best node, allowing the planner to operate in a ``best effort'' mode.

\subsubsection{Cost-to-Go Heuristic}
\label{heuristiccost}

\begin{figure}[!t]
\centering
    \includegraphics[width=0.8\columnwidth]{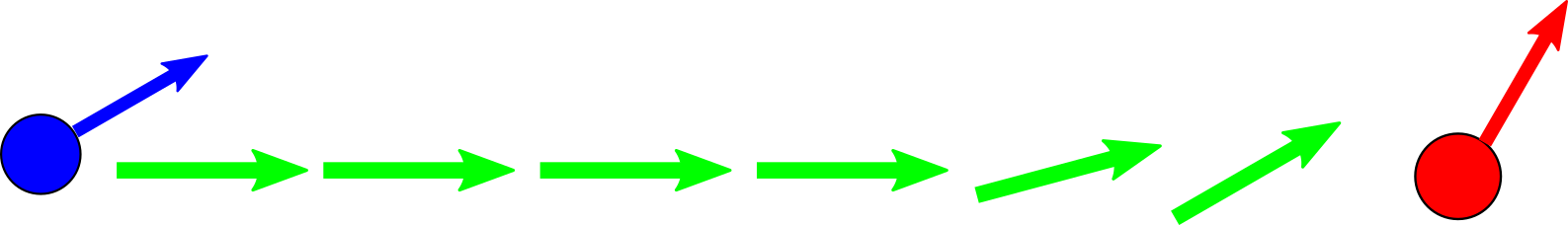}
\caption{
By introducing a heuristic cost estimate for the remaining path, the planning speed can be greatly increased. 
However, trying to achieve the nominal orientation of the end goal along the whole trajectory can result in the robot trying to walk sideways.
 Instead, we blend between the nominal heading pointing from the start (blue) to the goal (red) and the final end position to compute the nominal yaw setpoint.}
\label{fig:heuristic_cost_diagram}
\end{figure}

The selection of an appropriate heuristic cost-to-go estimate is critical to an efficient, fast search.
The role of the heuristic is to estimate the total remaining cost from each individual node to the goal, to predict the best remaining path.
We model our heuristic assuming the robot will walk straight towards the goal, and then start turning near the end.
We compute a reference desired orientation that changes accordingly, driving the robot towards the desired heading until it gets close to the goal, where it blends towards the final goal orientation.
This is shown by the arrows in \autoref{fig:heuristic_cost_diagram}, which point towards the goal until the end, where they start to match the goal yaw.
We then estimate the cost-to-go as the total Euclidean distance and rotation plus the minimum number of steps, multiplied by an inflation weight.

\subsection{Goal Evaluation and Anytime Planning}

If, during the expansion step, one of the child nodes is equal to either one of the goal nodes (either side), a valid path has been found.
If the heuristic cost has been inflated, this path is ultimately suboptimal.
Note that an ARA* planner would then continue to search by reducing the heuristic inflation and looking for an optimal solution.
While this is an area of interest for future work, we are more interested in finding fast, feasible, near-optimal solutions in this work, which can be provided with a low heuristic inflation.

If the search times out before a solution is found, the ``best effort'' path can be returned.
In this case, the best effort path is calculated as the path to the lowest cost node that is evaluated in the Edge Scoring step, and can be returned at any time.
This can be particularly useful when navigating in areas where dynamic obstacles may be present, so that the footstep plan can be continuously updated in response to these obstacles.
It also allows a plan to be found, even if the goal node is not feasible, such as being out of the robot's current line of sight.

\section{Edge Avoidance by Partial Foothold Wiggling}
\label{partialfootholdwiggling}

\begin{figure}[!b]
\vspace{-6mm}
\centering
    \includegraphics[width=0.7\columnwidth]{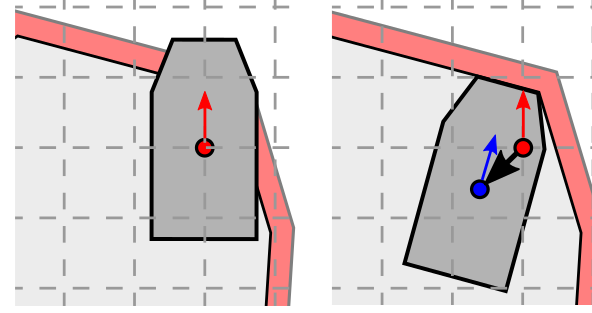}
\caption{The node placement when aligned with the grid may result in the foot being near the edge, or even hanging off the planar region, as shown on the left.
If possible, it is desired to move the node so that it is fully within the polygon by a certain distance.
This can be done by translating and rotating the foot polygon so that the node is at the blue dot on the right.}
\label{fig:wiggle_example}
\end{figure}

\begin{figure}[!b]
\centering
    \includegraphics[width=0.3\columnwidth]{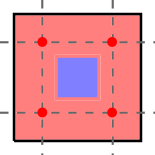}
\caption{If the footstep lattice lines up with a small planar region, it can happen so that none of the nodes in the lattice are within the region by the desired amount. 
In this example, the blue area is the area of the region that is that distance inside, while the red is not.
As can be seen, all the nodes, shown as the red dots, are within the red, unacceptable region.
Some method for shifting the nodes to the blue region is needed for this to be a valid step.}
\label{fig:discretization_error}
\end{figure}

One of the disadvantages of using a graph-search based planning approach is the risk that, due to the graph discretization, many possible solutions are ignored.
This becomes a particular problem when trying to avoid stepping near the edge of a planar region.
For many reasons, it is desirable to not step near the edge of regions, including tracking error, slip, and support shifting.
Additionally, when possible it is preferable to maximize the foothold area, rather than using a partial foothold.
For example, in \autoref{fig:wiggle_example}, we see that a foothold that aligns with the grid may be over the edge of a region. 
However, if this foothold is shifted between nodes on the lattice, it isn't as far inside as the obviously valid nodes, but is far enough inside to still be a very good choice.
If relying purely on the footstep graph, though, this would never be a considered solution.
As a more rare edge case, if we simply invalidate nodes that are too close to the edge, we may end up without any valid footstep nodes.
For example, see \autoref{fig:discretization_error}.
In this example, the area of good footholds is shown by the inner blue bound. 
As shown, there are no valid nodes that are within this blue region, with the four nodes in the planar region being the red circles.

Instead of invalidating nodes that are too close to the edge, we can choose to only invalidate nodes that don't have enough area, and shift them to be a desired distance inside the region in a post-processing step. 
This will allow nodes near the edge that could result in full footholds, like in \autoref{fig:wiggle_example}, and allow using area that would have no viable footholds otherwise, like in \autoref{fig:discretization_error}.
To do this post processing, we can set up an optimization problem for each node in the path where we constrain the foot to be a certain distance inside the region polygon while minimizing the total translation and rotation.
This has the effect of shifting the foot inside the polygon, as shown in \autoref{fig:wiggle_example}.

\begin{figure*}[t]
\centering
    \includegraphics[width=1.9\columnwidth]{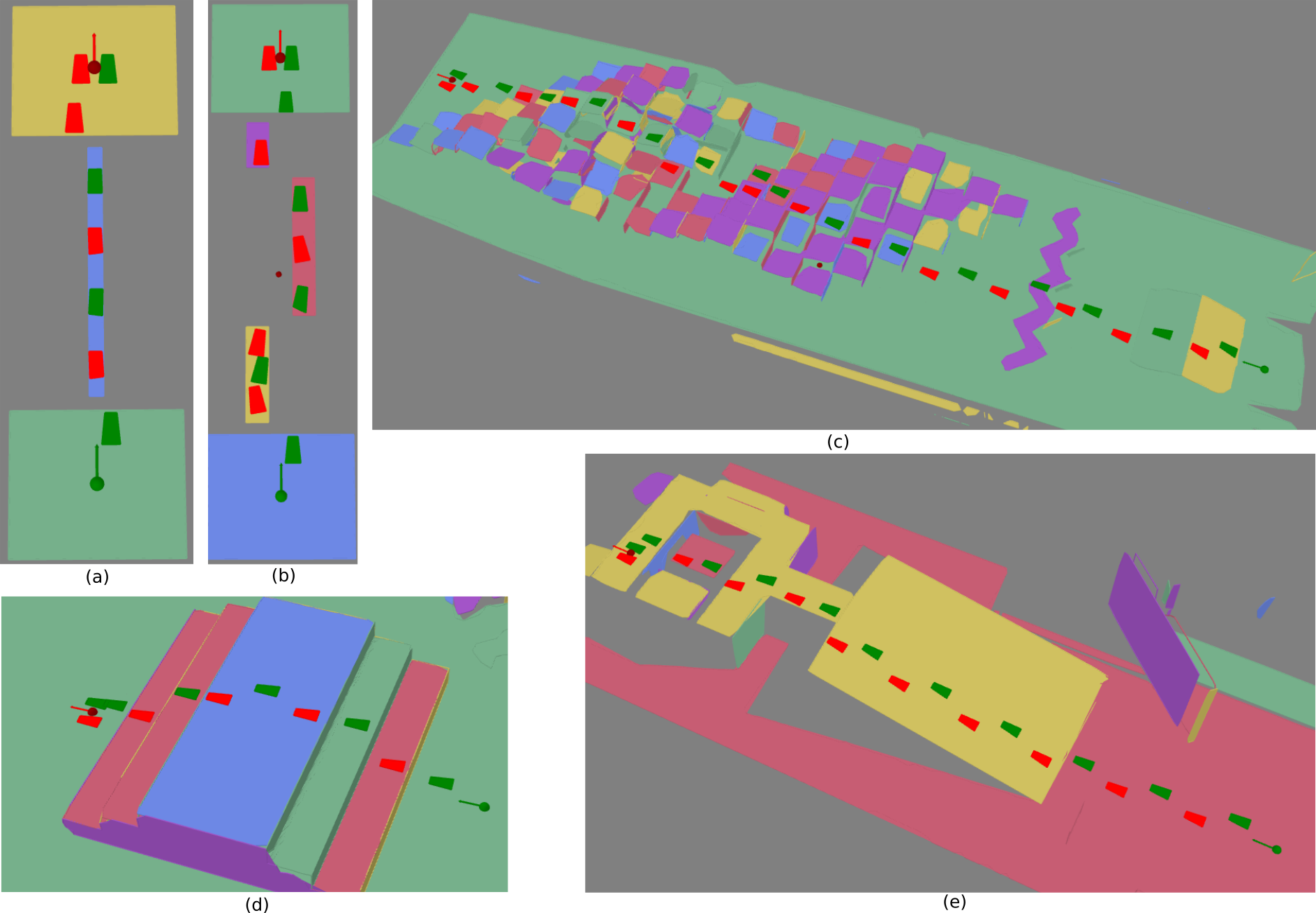}
\caption{Results of the footstep planner in several of our sample environments.
The start and goal positions are shown by the red and green spheres, respectively.
The planner is quickly able to determine the required step sequence to cross the environments.}
\label{fig:plan_results}
\vspace{-6mm}
\end{figure*}

To set up the optimization problem, we first define a constraint where all the points $\mathbf{x}_i \in \mathbb{R}^2$ that make up the footstep polygon are inside the polygon that defines the planar region, $\mathbf{P}_p$, by distance $d$.
This can be done by requiring the inequality constraint
\begin{equation}
\mathbf{A}_p \mathbf{x}_i \le \mathbf{b}_p(d), \ \forall i
\label{eqn:foothold_constraint}
\end{equation}
be satisfied, where $\mathbf{A}_p \in \mathbb{R}^{n \times 2}$ and $n$ is the number of edges in polygon $\mathbf{P}_p$.
The objective, then, is to find the a translation, $\mathbf{v}$, and rotation, $\theta$, of the step such that \autoref{eqn:foothold_constraint} holds.

If we define the centroid of the foot polygon as $\mathbf{r}_c$, the vector from the centroid to the $i^{th}$ foot polygon vertex is $\mathbf{r}_{p,i}$. 
Assuming that $\theta$ remains small, the location of the vertex after translation and rotation is then
\begin{equation}
\mathbf{x}_i = \mathbf{r}_c + \mathbf{r}_{p,i} + \mathbf{Jq} ,
\end{equation}
where
\begin{equation}
\mathbf{J} = 
\left[ \begin{array}{ccc}
1 & 0 & -v_{y,i} \\
0 & 1 & v_{x,i}
\end{array}\right], \ \ \ 
\mathbf{q}  = \left[ \mathbf{v}^T  \ \theta \right]^T.
\end{equation}
This allows us to then write our optimization problem as a quadratic program with 3 decision variables,
\begin{equation}
\begin{array}{rl}
\min_{\mathbf{q}} & \mathbf{q}^T \mathbf{Qq} \\ 
\text{subject to} & \mathbf{A}_p  \mathbf{Jq}   \le \mathbf{b}_p(d) - \mathbf{A}_p \left( \mathbf{r}_c + \mathbf{r}_{p,i} \right), \ \forall i \\
& \mathbf{q}_\text{min} \le \mathbf{q} \le \mathbf{q}_{\text{max}},
\end{array}
\end{equation}
where $\mathbf{Q}$ is a diagonal positive definite weighting matrix.

To keep this consistent with the node structure of the graph search, we must ensure that the desired distance inside $d$ is less than the discretization size. 
Otherwise, a different node should be chosen, such as selecting a node where the foothold would already be fully within the region in \autoref{fig:wiggle_example}.
That is, if achieving the desired distance inside the planar region requires moving a node past another valid node, that second node is the better option.

We execute this post-processing step on each of the nodes in the footstep planner solution.
This does mean that nodes can end up being inconsistent with the reachability limits imposed in \autoref{nodechecking}, as nodes on the bounds of the reachability can be shifted slightly outside of it.
However, as long as the maximum allowed shift distance is small (in our case, \SI{2}{\cm}), this is not of much concern.

\section{Results}
\label{results}

To aid in the development and testing of the footstep planner, we generated many 2D and 3D environments using planar regions, some of which are shown in \autoref{fig:plan_results}. 
For many of the 3D environments, such as \autoref{fig:plan_results}(c),(d), and (e), we ran a virtual LIDAR sensor over the terrain, generating planar regions using the approach outlined in \autoref{planarregionrepresentation}, to better represent actual robot data, which is imperfect, particularly in edge detection.
We then specified a start and goal pose of the robot, shown by the green and red spheres in \autoref{fig:plan_results}, respectively.
For the results presented in \autoref{fig:plan_results}  we use the same parameters that we use on the Atlas hardware in \autoref{fig:rough_terrain_results}.
As can be seen, this results in a variety of quality plans, with the resulting data being shown in \autoref{tab:plan_results}.
The times presented are generated using a desktop with a 4th gen 4-core i7 processor. 
As expected, as the number of required steps to reach the goal increases, the planning duration increases correspondingly.
It is worth noting that a large percentage of the nodes expanded on each iteration are found to be invalid, including 15\% rejected when planning on flat ground.
This suggests that if a smarter node expansion, such as that presented in \cite{Karkowski_2016b}, were used, the planning time could be correspondingly reduced.
One of the advantages, however, to having rough terrain where many nodes are found to be invalid is that the search-space of the problem is greatly reduced, leading to much faster results.

To illustrate the advantages of allowing partial footholds, we set up an environment consisting of two platforms spanned by a 4" beam, which was more narrow that the width of the foot, shown in \autoref{fig:plan_results}(a).
The reachability of the robot was also limited so that it could not put one foot directly in front of the other, requiring each step to be offset slight from the center of the beam.
If partial footholds were not allowed, the planner would not have been able to find solutions for \autoref{fig:plan_results}(b) either, which also required foothold cropping.

\begin{table}
\caption{Footstep Planner Results. This shows the statistics of the plans presented in \autoref{fig:plan_results}.}
\centering
\begin{tabular}{c c c c c c}
\hline \hline
Plan & Number & Plan & Planning & Nodes & Percent \\
& of Steps & Distance (m) & Duration (s) & Expanded & Rejected \\
\hline
(a) & 8 & 3.05 & 0.11 & 27 & 77.4 \\
(b) & 11 & 3.85 & 0.27 & 119 & 82.5 \\
(c) & 30 & 11.70 & 1.97 & 54 & 75.2 \\ 
(d) & 11 & 4.02 & 0.856 & 11 & 54.5 \\
(e) & 23 & 8.29 & 1.89 & 95 & 57.0
\label{tab:plan_results}
\end{tabular}
\end{table}

\begin{figure}[!t]
\centering
    \includegraphics[width=0.9\columnwidth]{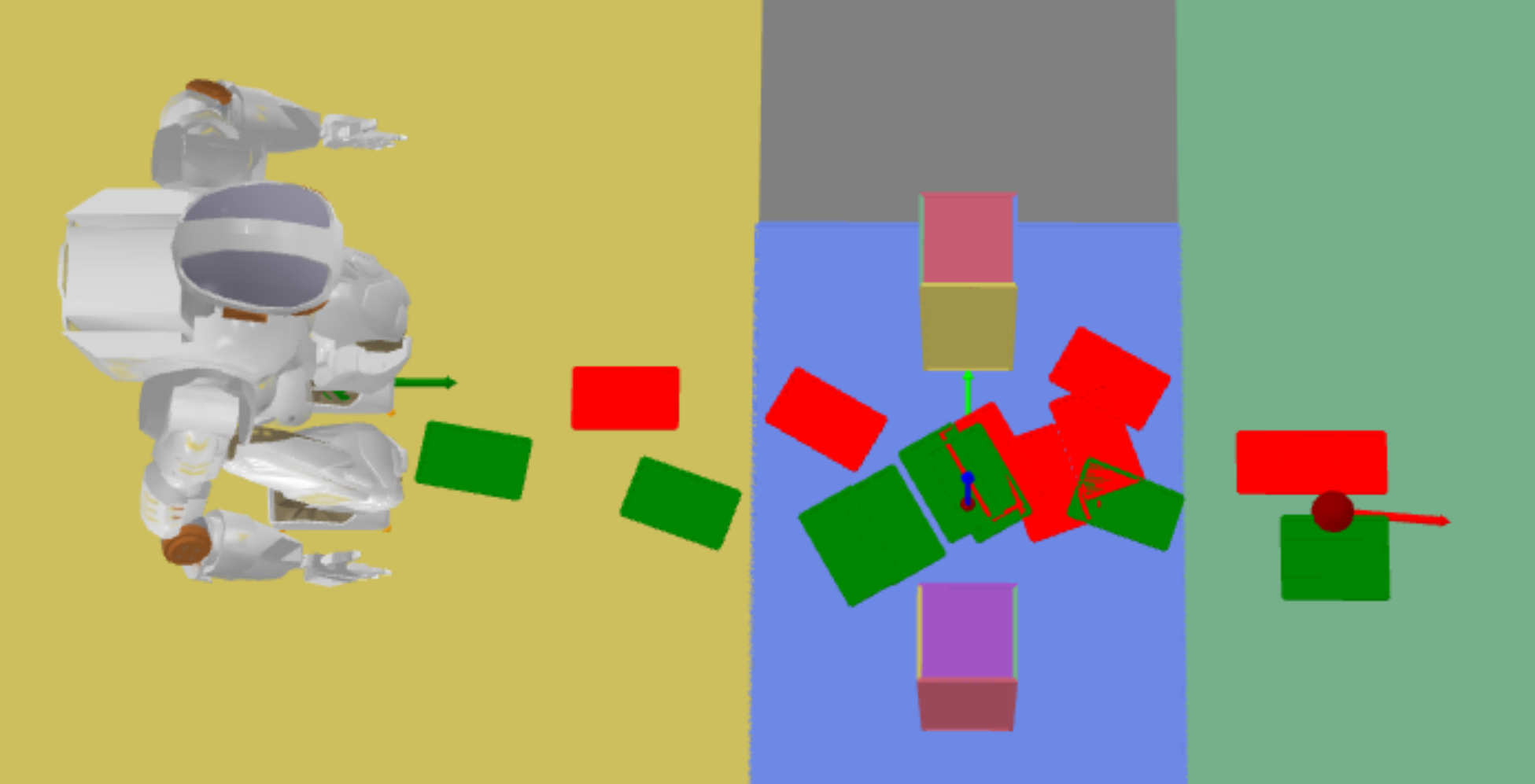}
\caption{Resulting plan of Valkyrie planning between a narrow gap, using sample data, where Valkyrie must turn sideways.}
\label{fig:between_bollards}
\end{figure}

We also explored the planner's capability at finding more advantages plans that require the robot to adjust its body using the bounding box collision model in \autoref{fig:obstacle_collision_checker} in \autoref{fig:between_bollards}.
In this environment, the robot had to turn sideways to squeeze between two vertical obstacles.
As expected, this requires a significantly longer planning time, expanding many nodes near the turn before the robot attempts to turn sideways, particularly if the heuristic inflation is high.

\begin{figure}[!t]
\centering
    \includegraphics[width=0.9\columnwidth]{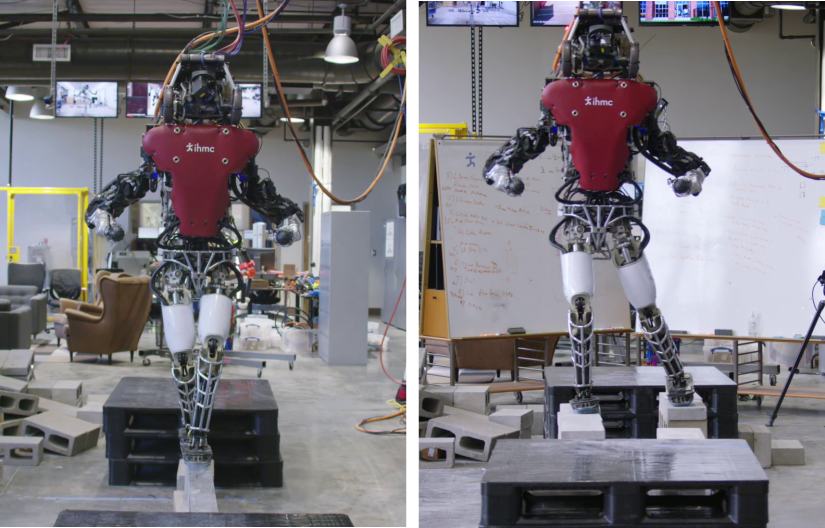}
\caption{Images of the robot executing several plans generated by the footstep planner. Here, the robot had to step with one foot almost directly in front of the other to cross the terrain. On the left, the minimum footstep area had to be lowered to 70\% for the robot to cross the environment. On the right, the footstep planner was adept at determining the proper sequence of steps to allow it to reach the other side.}
\label{fig:hardware_plan_results}
\end{figure}

We extensively tested this planner on hardware using Atlas and Valkyrie.
To model the environment, we used the approach briefly outlined in \autoref{planarregionrepresentation}. 
Several of the environments from \autoref{fig:plan_results} were directly reproduced for the robot, as shown in \autoref{fig:hardware_plan_results}.
For the results shown in \autoref{fig:rough_terrain_results}, we were able to achieve a relatively high success rate of approximately 90\%, with most failures due to balance errors, rather than planning errors.
One of the challenges that was noticed was the robot would often try and execute plans that would result in it stepping on its own feet, requiring the introduction and subsequent increase of the ``no-go'' area shown in \autoref{fig:node_checking}(c).
Additionally, to prevent the robot from taking steps that were both high or low and far forward or wide, we introduced a maximum length and width for steps that were above or below a certain distance. 
For \autoref{fig:hardware_plan_results}, our results were a little less reliable, achieving only about a 50\% success rate, although this was again primarily due to balance errors.
Additionally, for these examples, we had to reduce the required area for acceptance of the foothold to 70\%.

\begin{figure*}[!t]
\centering
    \includegraphics[width=1.9\columnwidth]{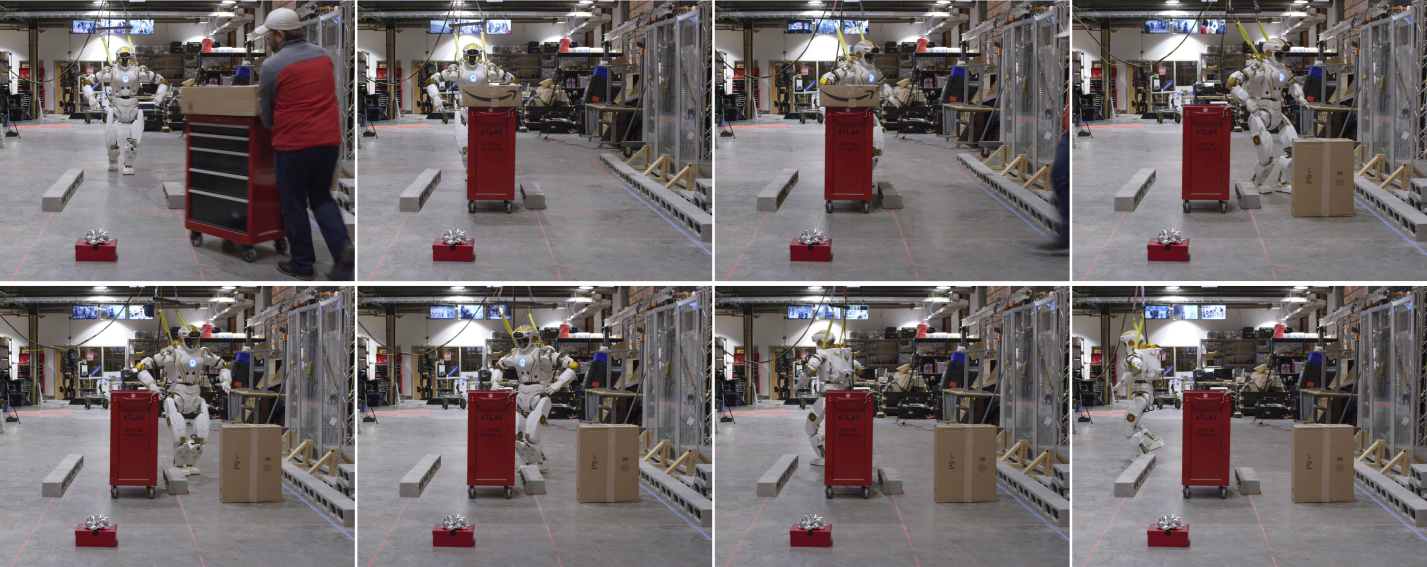}
\caption{Resulting plan of Valkyrie using the anytime capabilities of the footstep planner. The robot dynamically avoided objects placed in its path, replanning first to the left as it was a shorter distance, then to the right as it was the only remaining path.}
\label{fig:anytime_obstacle_avoidance}
\end{figure*}

Lastly, to test the anytime capabilities of the planner, we set up an environment where obstacles were put in between Valkyrie and the goal, requiring both their detection and rapid replanning to avoid collisions, as shown in \autoref{fig:anytime_obstacle_avoidance}.
In this experiment, the robot was told to plan to the package.
As it approached, an obstacle was placed in front of it, causing the robot to replan to its left.
Then another obstacle was placed in front of the robot, causing it to back up and plan to the right, eventually reaching the goal.

\section{Discussion and Future Work}
\label{discussion}

Several state-of-the-art approaches are moving towards a hierarchical/multi-stage planning approach, where the search-space of the low-level contact sequence planning is reduced through the use of a guiding heuristic plan or body path plan that can be found very quickly.
We believe that this type of hierarchy is not only an excellent method to increase planning speed, but is also more in line with nature, where the full contact sequence is not planned far in advance\citep{Matthis_2013}. 
We believe that by combining this type of short-horizon planning with our Weighted A* approach, we can get plans at a realtime rate, without having to resort to methods such as the adaptive action set described in\cite{Karkowski_2016b}, despite its excellent merits.
Additionally, environments such as that shown in \autoref{fig:between_bollards} can be challenging for A* search planners, as too high of a heuristic weight can cause the planner to get ``stuck'', only selecting nodes to expand that are close to the goal.
Indeed, the proverbial cul-de-sac problem can cause the planner to exhaustively check all the possible nodes in the cul-de-sac before continuing its search around.
We believe that the inclusion of a body path as a guiding heuristic will allow this type of environment to be solved very quickly, which address these type of problems well\citep{Karkowski_2016}.

One of the areas our approach is most lacking is a consideration of the CoM dynamics, as in\citep{Lin_2019}. 
While it is unclear if the learned-type dynamic cost prediction presented in that work is necessarily the best approach, it is certainly promising for producing plans that are dynamically variable to the environment structure.
It is likely that, as our algorithm is expanded to consider multi-contact motions, some kind of dynamic modeling will be required to weight between the contributions of hand vs. foot contacts.

We are also exploring modifying our existing Weighted A* approach to a full Anytime Repairing A* algorithm to continually improve the resulting plans once one is found\citep{Likhachev_2004}.

\subsection{Source Code and Media}

Our implementation of our planar region segmentation algorithm, our footstep planner, and our walking controller can be found on our GitHub page, \url{https://github.com/ihmcrobotics}.
The accompanying video can be found at \url{https://youtu.be/PqLZP8TANlg}.

\bibliography{mybib}

\begin{thebibliography}{10}
\providecommand{\url}[1]{#1}
\csname url@rmstyle\endcsname
\providecommand{\newblock}{\relax}
\providecommand{\bibinfo}[2]{#2}
\providecommand\BIBentrySTDinterwordspacing{\spaceskip=0pt\relax}
\providecommand\BIBentryALTinterwordstretchfactor{4}
\providecommand\BIBentryALTinterwordspacing{\spaceskip=\fontdimen2\font plus
\BIBentryALTinterwordstretchfactor\fontdimen3\font minus
  \fontdimen4\font\relax}
\providecommand\BIBforeignlanguage[2]{{%
\expandafter\ifx\csname l@#1\endcsname\relax
\typeout{** WARNING: IEEEtran.bst: No hyphenation pattern has been}%
\typeout{** loaded for the language `#1'. Using the pattern for}%
\typeout{** the default language instead.}%
\else
\language=\csname l@#1\endcsname
\fi
#2}}

\bibitem{Hornung_2012b}
A.~Hornung and M.~Bennewitz, ``Adaptive level-of-detail planning for efficient
  humanoid navigation,'' in \emph{2012 IEEE International Conference on
  Robotics and Automation (ICRA)}.\hskip 1em plus 0.5em minus 0.4em\relax IEEE,
  2012, pp. 997--1002.

\bibitem{Hornung_2012}
A.~Hornung, A.~Dornbush, M.~Likhachev, and M.~Bennewitz, ``Anytime search-based
  footstep planning with suboptimality bounds,'' in \emph{12th IEEE-RAS
  International Conference on Humanoid Robots (Humanoids)}, 2012, pp. 674--679.

\bibitem{Chestnutt_2003}
J.~Chestnutt, J.~Kuffner, K.~Nishiwaki, and S.~Kagami, ``Planning biped
  navigation strategies in complex environments,'' in \emph{3rd IEEE-RAS
  International Conference on Humanoid Robots (Humanoids)}, Oct 2003.

\bibitem{Chestnutt_2005}
J.~Chestnutt, M.~Lau, G.~Cheung, J.~Kuffner, J.~Hodgins, and T.~Kanade,
  ``Footstep planning for the {H}onda {A}simo humanoid,'' in \emph{2005 IEEE
  International Conference on Robotics and Automation (ICRA)}.\hskip 1em plus
  0.5em minus 0.4em\relax IEEE, 2005, pp. 629--634.

\bibitem{Chestnutt_2007}
J.~Chestnutt, K.~Nishiwaki, J.~Kuffner, and S.~Kagami, ``An adaptive action
  model for legged navigation planning,'' in \emph{2007 7th IEEE-RAS
  International Conference on Humanoid Robots (Humanoids)}.\hskip 1em plus
  0.5em minus 0.4em\relax IEEE, 2007, pp. 196--202.

\bibitem{Stumpf_2014}
A.~Stumpf, S.~Kohlbrecher, D.~C. Conner, and O.~von Stryk, ``Supervised
  footstep planning for humanoid robots in rough terrain tasks using a black
  box walking controller,'' in \emph{2014 IEEE-RAS International Conference on
  Humanoid Robots (Humanoids)}.\hskip 1em plus 0.5em minus 0.4em\relax IEEE,
  2014, pp. 287--294.

\bibitem{Deits_2014}
R.~Deits and R.~Tedrake, ``Footstep planning on uneven terrain with
  mixed-integer convex optimization,'' in \emph{14th IEEE-RAS International
  Conference on Humanoid Robots (Humanoids)}, 2014, pp. 279--286.

\bibitem{Karkowski_2016}
P.~Karkowski and M.~Bennewitz, ``Real-time footstep planning using a geometric
  approach,'' in \emph{2016 IEEE International Conference on Robotics and
  Automation (ICRA)}.\hskip 1em plus 0.5em minus 0.4em\relax IEEE, 2016, pp.
  1782--1787.

\bibitem{Lin_2017}
Y.-C. Lin and D.~Berenson, ``Humanoid navigation in uneven terrain using
  learned estimates of traversability,'' in \emph{2017 IEEE-RAS 17th
  International Conference on Humanoid Robotics (Humanoids)}.\hskip 1em plus
  0.5em minus 0.4em\relax IEEE, 2017, pp. 9--16.

\bibitem{Lin_2019}
Y.-C. Lin, B.~Ponton, L.~Righetti, and D.~Berenson, ``Efficient humanoid
  contact planning using learned centroidal dynamics prediction,'' \emph{arXiv
  preprint arXiv:1810.13082}, 2018.

\bibitem{Karkowski_2016b}
P.~Karkowski, S.~O{\ss}wald, and M.~Bennewitz, ``Real-time footstep planning in
  3d environments,'' in \emph{2016 IEEE-RAS 16th International Conference on
  Humanoid Robots (Humanoids)}.\hskip 1em plus 0.5em minus 0.4em\relax IEEE,
  2016, pp. 69--74.

\bibitem{Ratliff_2009}
N.~Ratliff, M.~Zucker, J.~A. Bagnell, and S.~Srinivasa, ``{CHOMP}: Gradient
  optimization techniques for efficient motion planning,'' in \emph{2009 IEEE
  International Conference on Robotics and Automation (ICRA)}.\hskip 1em plus
  0.5em minus 0.4em\relax IEEE, 2009.

\bibitem{Winkler_2018}
A.~W. Winkler, C.~D. Bellicoso, M.~Hutter, and J.~Buchli, ``Gait and trajectory
  optimization for legged systems through phase-based end-effector
  parameterization,'' \emph{IEEE Robotics and Automation Letters}, vol.~3,
  no.~3, pp. 1560--1567, 2018.

\bibitem{Ponton_2016}
B.~Ponton, A.~Herzog, S.~Schaal, and L.~Righetti, ``A convex model of humanoid
  momentum dynamics for multi-contact motion generation,'' in \emph{2016
  IEEE-RAS 16th International Conference on Humanoid Robots (Humanoids)}.\hskip
  1em plus 0.5em minus 0.4em\relax IEEE, 2016, pp. 842--849.

\bibitem{Herdt_2010}
A.~Herdt, H.~Diedam, P.-B. Wieber, D.~Dimitrov, K.~Mombaur, and M.~Diehl,
  ``Online walking motion generation with automatic footstep placement,''
  \emph{Advanced Robotics}, vol.~24, no. 5-6, pp. 719--737, 2010.

\bibitem{Naveau_2016}
M.~Naveau, M.~Kudruss, O.~Stasse, C.~Kirches, K.~Mombaur, and P.~Sou{\`e}res,
  ``A reactive walking pattern generator based on nonlinear model predictive
  control,'' \emph{IEEE Robotics and Automation Letters}, vol.~2, no.~1, pp.
  10--17, 2016.

\bibitem{Carpentier_2016}
J.~Carpentier, S.~Tonneau, M.~Naveau, O.~Stasse, and N.~Mansard, ``A versatile
  and efficient pattern generator for generalized legged locomotion,'' in
  \emph{2016 IEEE International Conference on Robotics and Automation
  (ICRA)}.\hskip 1em plus 0.5em minus 0.4em\relax IEEE, 2016, pp. 3555--3561.

\bibitem{Ebendt_2009}
R.~Ebendt and R.~Drechsler, ``Weighted {A}* search-unifying view and
  application,'' \emph{Artificial Intelligence}, vol. 173, no.~14, pp.
  1310--1342, 2009.

\bibitem{Deits_2015}
R.~Deits and R.~Tedrake, ``Computing large convex regions of obstacle-free
  space through semidefinite programming,'' in \emph{Algorithmic foundations of
  robotics XI}.\hskip 1em plus 0.5em minus 0.4em\relax Springer, 2015, pp.
  109--124.

\bibitem{Gutmann_2008}
J.-S. Gutmann, M.~Fukuchi, and M.~Fujita, ``3d perception and environment map
  generation for humanoid robot navigation,'' \emph{The International Journal
  of Robotics Research}, vol.~27, no.~10, pp. 1117--1134, 2008.

\bibitem{Matthis_2013}
J.~S. Matthis and B.~R. Fajen, ``Humans exploit the biomechanics of bipedal
  gait during visually guided walking over complex terrain,'' \emph{Proceedings
  of the Royal Society B: Biological Sciences}, vol. 280, no. 1762, p.
  20130700, 2013.

\bibitem{Likhachev_2004}
M.~Likhachev, G.~J. Gordon, and S.~Thrun, ``{ARA}*: Anytime {A}* with provable
  bounds on sub-optimality,'' in \emph{Advances in neural information
  processing systems}, 2004, pp. 767--774.

\end{thebibliography}

\end{document}